# Uncovering Conspiratorial Narratives within Arabic Online Content

**Djamila Mohdeb**
University of Mohamed Seddik Benyahia
Jijel, Algeria
Laboratory of Intelligent Systems and Cognitive Computing (ISCC)
Bordj Bou Arreridj, Algeria
djamila.mohdeb@univ-jijel.dz

**Meriem Laifa**
University of Mohamed Bachir El Ibrahimi
Bordj Bou Arreridj, Algeria
Laboratory of Informatics and its Applications of M'sila (LIAM)
M'sila, Algeria
meriem.laifa@univ-bba.dz

**Zineb Guemraoui and Dalila Behih**
University of Mohamed Bachir El Ibrahimi
Bordj Bou Arreridj, Algeria
{zineb.guemraoui0, dalilabehih}@gmail.com

## Abstract

This study investigates the spread of conspiracy theories in Arabic digital spaces through computational analysis of online content. By combining Named Entity Recognition and Topic Modeling techniques, specifically the Top2Vec algorithm, we analyze data from Arabic blogs and Facebook to identify and classify conspiratorial narratives. Our analysis uncovers six distinct categories: gender/feminist, geopolitical, government cover-ups, apocalyptic, Judeo-Masonic, and geoengineering. The research highlights how these narratives are deeply embedded in Arabic social media discourse, shaped by regional historical, cultural, and sociopolitical contexts. By applying advanced Natural Language Processing methods to Arabic content, this study addresses a gap in conspiracy theory research, which has traditionally focused on English-language content or offline data. The findings provide new insights into the manifestation and evolution of conspiracy theories in Arabic digital spaces, enhancing our understanding of their role in shaping public discourse in the Arab world.

*K*eywords Conspiracy theories · Misinformation · Arabic Natural Language Processing · Topic Modeling · Named Entity Recognition · Social Networks · Digital Environement

## 1 Introduction

Conspiracies, defined as covert schemes orchestrated by groups for illicit or harmful purposes [1], hold a distinct psychological appeal. They tap into the fundamental human need for safety and security, fostering vigilance against perceived threats. This stems from the belief that understanding one's circumstances can help avert future adversities and control one's destiny [2].

While conspiracies are verifiable incidents, conspiracy theories are unproven, potentially false perceptions[3]. These theories, often categorized as misinformation [4], attribute major events to secret plots by powerful groups acting against the common good [3].

Conspiracy theorists may not intentionally misinform; many genuinely fear conspiracies or enjoy speculative explanations [5]. Nevertheless, these narratives can have far-reaching consequences, influencing public discourse, social cohesion, and political stability [6]. For instance, QAnon conspiracy theory has been linked to violence and political influence [7], while anti-vaccination theories have affected public health efforts[8].

Conspiracy theorizing has been a longstanding phenomenon throughout history [9]. However, its visibility has greatly intensified, making it an undeniable issue in contemporary times. This escalation can be attributed primarily to the rapid growth of digital information and the widespread use of online social media platforms [10]. Indeed, within the realm of the digital landscape, characterized by its wealth of information, conspiracy theories assume a prominent role. They serve as valuable instruments for constructing coherent narratives by connecting fragmented pieces of information [11]. These narratives empower individuals to derive meaning from the overwhelming influx of data and discern meaningful patterns from what might initially appear as a chaotic assortment of isolated facts [11].

While extensive research has explored conspiracy theories online, most studies focus on English-language content and Western nations [12].The Arab context presents a distinct case, with conspiracist thinking becoming increasingly prominent in Arabic-speaking online communities. The abundance and evolving nature of this content necessitate automated approaches for identification and classification.

Natural Language Processing (NLP) methodologies, particularly Topic Modeling (TM) and Named Entity Recognition (NER), can be valuable in uncovering conspiratorial content [13][14]. However, their application to Arabic conspiracist content remains largely unexplored. This research aims to bridge this gap by combining NER and TM to automatically uncover conspiracy theory topics and narratives in Arabic-language texts from Facebook and online blogs. We will discern underlying themes, examine their historical and cultural sources, and analyze named entities to provide insights into the actors involved in these narratives.

Our findings can deepen understanding of conspiracy theories, contribute to countering misinformation, and aid researchers, journalists, and policymakers in addressing this societal challenge.

The remaining sections of this paper adhere to the following structure. Following the literature review, we elaborate on the methodology employed in this study. Subsequently, we present the experimental setup then the findings. The paper concludes with a comprehensive discussion of the results and conclusions, encompassing an analysis of the theoretical and practical implications derived from our discoveries, as well as an exploration of the limitations and potential directions for future research.

## 2 Literature Review

### 2.1 Conspiracy Theories in Arabic-speaking Sphere

Despite the prevalence of conspiracy theories, researchers from psychological, social, and political science disciplines have only recently started studying them in depth, particularly within the past two decades [15]. Interestingly, a significant portion of these studies have been motivated by the urgent need to address the challenges posed by the COVID-19 public health crisis [16]. In fact, it is noteworthy that more than half of the academic publications on conspiracy theories in psychology have been published since 2019, reflecting the heightened attention on this topic in light of recent events [12].

The body of literature on conspiracism in Arab-speaking countries, specifically the Middle East and North Africa, is notably limited. Two early scholarly works, namely [17] and [18], primarily adopted a pathological approach (i.e. approach that views conspiracism as a psychological or sociological disorder), which was prominent in the initial theoretical investigations of conspiracism. While the former study played a pivotal role in initiating the discourse on conspiracism in the Middle East, the latter aimed to establish a link between conspiracy thinking in this region and theories concerning paranoid processes (paranoid processes refer to patterns of thought and behavior characterized by extreme suspicion, mistrust, and a belief that others are conspiring or plotting against an individual or group).

Moving beyond the pathological approach, researcher Matthew Gray [19]in his seminal paper addressed the inadequacy of previous studies in understanding conspiracy theories within the Middle Eastern context. He emphasized the significance of considering the regions unique cultural, political, and historical characteristics when examining conspiracism. Gray critically examined the applicability of existing theories to non-Western and specifically Arab Middle Eastern societies, highlighting both the potential areas of relevance and the distinct features that differentiate the Middle East. Meanwhile, researchers involved in a comparative collective study conducted by Butter and Reinkowski in 2014 [20] shed light on the multifaceted aspects of conspiracy theories and their societal implications in both the USA and the Middle East, utilizing a variety of theoretical and methodological perspectives. Their research provided valuable insights into the nature and function of conspiracy theories, highlighting their origins in feelings of powerlessness (i.e. feelings of lacking control over one's own life) and the desire to understand complex phenomena. Furthermore, these scholars argued that conspiracy theories can be used to rationalize acts of violence and serve as a tool to advance political agendas.

Another research undertaken by [21] explored the association between anti-Western and anti-Jewish attitudes, political knowledge, education, and belief in conspiracy theories and misperceptions. The findings revealed widespread adherence to conspiracy theories, a strong association between conspiracy beliefs and anti-Western/anti-Jewish attitudes, and no evidence supporting the powerlessness hypothesis regarding the influence of feelings of control on conspiracy beliefs . Likewise, the work of [22] emphasized that sectarianism and exposure to foreign intervention in violent conflict are the primary factors that explain trust in United States -related conspiracy theories in the Arab Middle East.

## 2.2 Conspiracy Theories in Online Environments

The rise of online environments and the widespread use of social media platforms have prompted extensive research on conspiracy theories [12]. As individuals increasingly rely on online platforms for information and social interactions, conspiracy theories have gained notable visibility and influence. Social media, in particular, facilitates the rapid dissemination of conspiratorial narratives, reaching large audiences expeditiously [23]. This has prompted interdisciplinary research endeavors, encompassing psychology, sociology, communication studies, and computer science, to comprehend the psychological drivers behind belief in conspiracy theories, analyze their social dynamics, and develop computational methods for detection and analysis [12].

Various methodological approaches have been employed to analyze conspiratorial narratives on the web and social media platforms. These include content analysis [24][25], social network analysis [23], sentiment analysis [26], qualitative discourse analysis [27], and computational techniques such as topic modeling [28] and machine learning [29].
Nevertheless, a recent survey conducted by [12] has identified empirical biases in conspiracy theory research, including a focus on mainstream social media platforms, Western countries, English-language communication, and specific conspiracy theory topics. Accordingly, only a few studies have specifically investigated conspiracy theories in Arabic online content. For instance, [30] utilized a textual analysis approach that leverages linguistic cues to investigate the conceptualization of geopolitical concerns within a substantial corpus of Egyptian tweets. Their research revealed that conspiratorial thinking is pervasive, often serving as a coping mechanism to mitigate growing anxiety arising from the rapidly evolving political landscape in Egypt and the broader region. Similarly, in another study conducted by [22], a sentiment analysis approach was applied to Arabic Twitter posts spanning from 2014 to 2015. The research revealed a significant negative perception of the United States within the Middle East region. This negative sentiment led to citizens forming conspiratorial associations between the United States and the Islamic State of Iraq and the Levant (ISIS).
Moreover, it is noteworthy that a substantial portion of research concerning conspiratorial narratives within Arabic online communities has primarily concentrated on conspiracy theories pertaining to the COVID-19 health crisis. However, these studies have typically not prioritized conspiracy theories as their central focus but have rather provided brief examinations of such theories in conjunction with other manifestations of inaccurate information [31] [25].

In response to the aforementioned literature gaps, our study aims to provide an in-depth exploration of the landscape of conspiracy thinking within the Arabic online sphere. Specifically, we seek to illuminate the most widely debated conspiracy theory topics by employing a combined approach that integrates TM and NER. We selected Facebook and the online Arabic blogosphere as our data sources, as the majority of existing studies have primarily focused on offline data collection methods such as surveys and questionnaires, or they have centered on Twitter data.

Through this comprehensive exploration, we aspire for our study to serve as a good starting point for a more nuanced understanding of conspiracy theories within the Arabic online landscape, while addressing the current gaps in the literature and expanding knowledge on this topic.

## 3 Data Exploration

### 3.1 Data Collection

To construct a comprehensive dataset concerning conspiracy theories, we amassed a total of 2,048 Arabic texts composed in both Modern Standard Arabic (MSA) and dialectal Arabic. These texts were drawn from various sources, encompassing public pages and groups on Facebook [1], along with an Arabic online blog [2] and covering a wide range of subjects, including politics, science, economics, society, entertainment, and culture. Our data collection, spanning from June 2010 to March 2023, was designed to incorporate diverse perspectives and thematic aspects, thereby providing an extensive overview of the subject matter.

A meticulous curation process was subsequently employed to refine the dataset. This process entailed the exclusion of non-informative documents, exceedingly brief documents (akin to texts comprising fewer than five words), duplicated documents, and documents featuring only URLs and symbols. Moreover, Arabic-specific preprocessing techniques were applied. These techniques involved normalizing Arabic words, eliminating Arabic diacritics and punctuation marks, as well as filtering out stop words. This rigorous preprocessing served to enhance the overall quality of the data.

Importantly, our investigation focuses exclusively on textual content, thus disregarding multimedia components like images and videos. Following these careful and systematic steps, we successfully compiled a conclusive corpus which we named ArCons Dataset, encompassing 1,641 Arabic documents centered around conspiracy theories. A summarized depiction of the ArCons dataset's attributes is provided in Table 1 .

Table 1: ArCons Dataset Description

| Dataset | **ArCons Dataset** |
|---|---|
| Source | Social Media (Facebook), Blogs |
| Timespan | 2010 to 2023 |
| Total number of samples | 1 641 documents |
| Type of samples | Textual |
| Language of documents | Modern Standard Arabic (MSA), Dialectal Arabic (Algerian, Egyptian, Moroccan, Syrian) |
| Minimum document length | 07 words |
| Maximum document length | 4 405 words |
| Average document length | 316.93 words |
| Total number of words | 520 084 words |

### 3.2 Data Distribution

The corpus comprises 1,641 Arabic documents, of which 24% (392 documents) are extracted from a blog and the remaining 76% (1,249 documents) are sourced from Facebook.

The lengths of the documents within the corpus exhibit a diverse distribution. The histogram of the length of the documents in Figure 1 underscores a prevalent trend: Most documents have a length less than 1,000 words. As the lengths of the documents surpass this threshold, there is a noticeable decline in frequency, a pattern that continues as the lengths extend further. A smaller fraction of documents exhibit lengths exceeding 3000 words.

---

[1]https://www.facebook.com

[2]http://lsaidbiasl.blogspot.com/

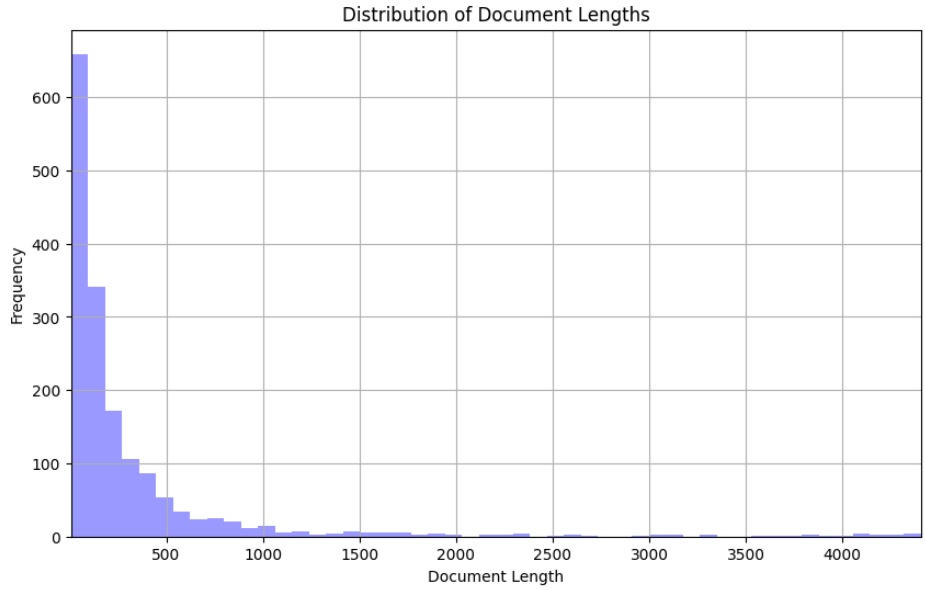

Figure 1: Distribution of documents' lengths in the corpus

## 3.3 Temporal Analysis

Table 2: Most frequent terms in the dataset

| Data | Frequent terms |
|------|----------------|
| First subset [2010–2016] | المسيح، العرب، الاوسط، رئيس، المنطقة، اليهود، الشعب، الكبرى، الثورة، الصهيونية، الدولة، بريطانيا، الملك، الشيطان، الدولار، عالمية، العين، المؤامرة، بوش، الصين، الحقيقة، الطاقة، الاقتصاد، الجيش، الاعلام، الخليج، الحروب، اوروبا، القاعدة، السعودية، السرية، العسكرية، الفوضى، المال، روسيا، تاريخ، السيطرة، روتشيلد، الذهب، الدين، النفط، البنوك، سوريا |
| | English translation: Christ, Arabs, Middle East, President, region, Jews, people, major, revolution, Zionism, state, Britain, king, devil, dollar, international, eye, conspiracy, Bush, China, truth, energy, economy, army, media, Gulf, wars, Europe, Al-Qaeda, Saudi Arabia, secrecy, military, chaos, money, Russia, history, control, Rothschild, gold, religion, oil, banks, Syria. |
| Second subset [2017–2023] | النساء، الرجل، السماء، الشمس، الحياة، الدولة، الحقيقة، الطاقة، الدين، القران، الشيطان، الشعب، تركيا، العلم، النسوية، الكون، الجيش، الفضاء، المجتمع، النبي، البشرية، العرب، الجن، ناسا، وهم، التاريخ، الملك، العلماء، القمر، نظرية، الحق، النار، البحر، الغرب |
| | English translation: Women, man, sky, sun, life, state, truth, energy, religion, Quran, devil, people, Turkey, knowledge, feminism, universe, army, space, society, prophet, humanity, Arabs, Jinn, NASA, illusion, history, king, scientists, moon, theory, justice, fire, sea, west. |

In order to examine and analyze the evolving interests of Arabic conspiracy theorists on the digital landscape, we partitioned the dataset into two distinct temporal categories: information spanning the interval from 2010 to 2016, and data encompassing the years 2017 to 2023. Table 2 presents the most frequent terms observed in two distinct time periods of the collected data. Our key observations from the two subsets of the data include:

- There appears to be a shift in the dominant topics of interest between the two time periods. The first part predominantly centers around geopolitical matters, power dynamics, and international players. The second part, however, emphasizes more diverse and contemplative topics, including humanity, societal/gender roles, spirituality, science and cosmic elements.

- Religious concepts and spiritual elements are prominent in both periods, but their incorporation in the second part is more nuanced and integrated into broader discussions.

The contrast in the frequent terms between the two time periods may reflect changing societal interests, concerns, and priorities. The transition from geopolitically oriented discussions to more introspective and comprehensive thematic inquiries suggests a potential evolution in prevailing discourse over time. These shifts can be attributed to a multitude of factors, and the subsequent considerations underscore the following salient points.

It is imperative to highlight that the period preceding 2019 was notably characterized by substantial political instability in various Arabic countries, notably ascribed to the Arab Spring uprisings. Consequently, it is unsurprising to observe the first segment of the data being preoccupied with geopolitics. In contrast, the emergence of the Covid-19 health crisis in post-2019 era diverted attention from the prevailing political and security landscape in the region. This led to critical reflections regarding the integrity of scientific enterprise and findings, thereby revitalizing conspiracy theories that cast doubt upon scientific accomplishments [32] [33]. Furthermore, the global upsurge in gender equality movement has significantly fueled conspiratorial narratives pertaining to societal structures and gender roles [34].

While our data does not fully depict Arabic conspiracy theorists web interests, the observations highlight the interplay between global events and conspiracy discussions, offering an approximate view of evolving interests of Arab conspiracy theorists over time.

## 3.4 Tracing Conspiracy Threads: NER Insights

Named Entity Recognition is the task of automatically identifying named entities in text and classifying them into predefined categories such as person names, organizations, locations, time, events and so on [14]. In our context, NER can be regarded as a valuable tool for delineating the intricate facets of Arabic conspiratorial thinking. The extracted entities, when enriched with context, can seamlessly complement the insights derived from Topic Modeling algorithms, thereby enhancing their depth and accessibility.

By employing the pre-trained Marifa NER model [3] designed for the Arabic language, we extracted the most frequently occurring entities from the compiled dataset across five distinct categories: locations, nationalities, persons, jobs, organizations, events, and products. Table 3 provides illuminating perspectives on the named entities that commonly intertwine with conspiracy theories within the realm of Arabic online content.

Overall, the extracted named entities underscore a significant focus on geopolitical dynamics, historical and contemporary figures, as well as diverse organizations, all framed within the context of conspiracy theories. It is intriguing to observe that numerous entities falling within the Location and Nationality categories pertain to countries or nationalities that have been entangled in geopolitical conflicts within the MENA region -and beyond- over recent years. Concurrently, the inclusion of notable names like Hitler, Obama, Sisi, Mohammed, Putin, and others implies a fascination with both past and present figures. These figures are often perceived as exerting influence in these contemporary conflicts, alongside other global disputes, or potentially possessing a degree of control over such occurrences. The trajectory of these insights finds reinforcement in the diverse entities identified across the remaining categories.

Entities featured in the Event category reflect a disposition towards skepticism and unease concerning a range of events, including national and international political upheavals (e.g. Arab Spring and Burma massacres), economic downturns (e.g. Euro collapse), health crises (e.g. Swine flu), and environmental catastrophes such as seismic events (e.g. Turkey earthquake). Evidently, the entities within the Organization category are further segregated into political entities (e.g. Congress), military entities (e.g.NATO), media outlets (e.g.Al Jazeera Channel and New York Times), and secret societies (e.g. Freemasons and the Illuminati). These entities are often implicated in the genesis of these crises, with allegations of orchestrating sinister schemes facilitated by potent mechanisms and weaponry, itemized in the Product category. The occupations outlined

---

within the Job category further corroborate this interpretation, as the extensive spectrum encompasses a multitude of professions, prominently featuring political and media roles.

The extraction of named entities outlines the contours of the corpus content. A more profound analysis of the context and substance of the documents, utilizing Topic Modeling algorithms, could provide supplementary insights into the prevalent narratives and themes within the corpus.

Table 3: Named entities within the dataset

| Category | Entities |
| --- | --- |
| Location | بريطانيا، قطر، روسيا، اوروبا، تركيا، واشنطن، مضيق هرمز، الصين، سوريا، السودان، ليبيا، السعودية، القدس <br> English translation: Britain, Qatar, Russia, Europe, Turkey, Washington, Strait of Hormuz, China, Syria, Sudan, Libya, Saudi Arabia, Jerusalem. |
| Nationality | الروسي، المصري، الايراني، الامريكية، البريطانية، الصهيونية، السعودية، الصيني، الاسرائيل، الالمانية، الفرنسي، القطري، اليهودية <br> English translation: Russian, Egyptian, Iranian, American, British, Zionist, Saudi Arabian, Chinese, Israeli, German, French, Qatari, Jewish |
| Person | المسيح، هتلر، اوباما، السيسي، محمد، بوتين، موسى، راسبوتين، فرعون، روتشيلد، بوش، كيسنجر، صدام، سليمان <br> English translation: Jesus Christ, Hitler, Obama, Sisi (Current President of Egypt), Mohammed (Prophet of Islam), Putin (Russian President), Moses (Prophet of Judaism), Rasputin (Russian mystic), Pharaoh (Egyptian ruler), Rothschild (Banking family), Bush (American President), Kissinger (American diplomat), Saddam (Iraqi President), Solomon (Prophet of Judaism) |
| Event | زلزال تركيا، اولمبياد لندن، الربيع العربي، الاقتصاد الاوروبي، زلزال اليابان، مشروع هارب، الحرب العالمية الثالثة النووية، مجازر العرب، مجازر بورما، مخطط زرع الفتنة، الانتفاضة المصرية، انهيار اليورو، اضطرابات اوروبا، اغتيال بن لادن، الانهيار الاقتصادي القادم، الاعلان الصهيوني، انفلونزا الخنازير <br> English translation: Turkey earthquake, London Olympics, Arab Spring, European economy, Japan earthquake, Project HAARP, Third World nuclear War, Arab massacres, Burma massacres, Plotting sedition, Egyptian uprising, Euro collapse, European unrest, Bin Laden assassination, Coming economic collapse, Zionist declaration, Swine flu. |
| Organization | الاتحاد الاوروبي، الجيش المصري، البنك الدولي، ناسا، الناتو، الشرطة البريطانية، الصهيونية، الكيان الصهيوني، قناة الجزيرة، الكونغرس، الامبراطورية الصهيونية العالمية، الماسونيين، الاتحاد السوفياتي، البنتاغون، الجيش الاسرائيلي، الدولة العثمانية، بنك التسويات الدولية، الاخوان المسلمون، نيويورك تايمز، المتنورون <br> English translation: The European Union, the Egyptian Army, the World Bank, NASA, NATO, the British Police, Zionism, the Zionist entity, Al Jazeera Channel, Congress, the Global Zionist Empire, the Freemasons, the Soviet Union, the Pentagon, the Israeli Army, the Ottoman Empire, the International Settlement Bank, the Muslim Brotherhood, The New York Times, the Illuminati. |
| Job | وزير الملك، رئيس الوزراء، رئيس، الجنرال، الملكة، الادميرال، وزير المالية، السفير، العلماء، وزير الخارجية، امير، حاخام، الدكتور، الصحفي، الكاتب <br> English translation: Minister, king, prime minister, president, general, queen, admiral, finance minister, ambassador, scientists, foreign minister, prince, rabbi, doctor, journalist, writer. |
| Product | النفط، سلاح هارب، الكرت الماسوني، خبز، الذهب، الاقمار الصناعية، الماء، الاسلحة الكهرومغناطيسية، الصواريخ، التيليجرام، القنوات الفضائية، القنبلة، الكيمتريل، المكياج <br> English translation: Oil, HAARP weapon, Masonic card, bread, gold, satellites, water, electromagnetic weapons, missiles, Telegram, satellite channels, bomb, Chemtrails, makeup |

## 3.5 Topic Modeling for detecting conspirasionist content

Topic modeling is a computational technique used to uncover latent thematic structures within textual data collections. It organizes and analyzes text based on shared patterns of word co-occurrence, revealing coherent themes that capture the underlying semantic content of documents [13].

In this study, we employed a range of topic modeling algorithms to extract themes from our corpus. These algorithms included both classical and advanced techniques. The classical methods comprised Latent Dirichlet Allocation (LDA) [35], Non-Negative Matrix Factorization (NMF) [36], and Latent Semantic Analysis (LSA) [37]. The advanced methods included BERTopic [38], Top2Vec [39], and Contextualized Topic Model (CTM) [40].

While a comprehensive comparison of these algorithms' performance lies beyond the scope of this study, Top2Vec emerged as the most effective for our purposes, consistently producing the most coherent topic results.

Table 4 presents the topics extracted using Top2Vec. Each extracted topic underwent careful examination to assess its potential relevance to conspiracy theories. It is crucial to note that while these topics may provide fertile ground for conspiracy theories, not all discussions within these topics are necessarily conspiratorial in nature.

Our analysis reveals that the predominant subjects explored by Arabic-speaking conspiracy theorists on the Web encompass the following themes:

- **Topic 01:** This topic seems to revolve around discussions related to social dynamics, gender roles, feminism, and societal norms.

- **Topic 02:** This topic is likely concerned with discussions related to international politics, military actions, and regional conflicts.

- **Topic 03:** This topic appears to be focused on discussions related to space exploration, extraterrestrial life, and UFOs.

- **Topic 04:** This topic seems to involve discussions about cosmological concepts, the shape of the Earth, and references from religious texts.

- **Topic 05:** This topic is likely centered around religious discussions regarding the end times, prophecies, and figures of spiritual significance.

- **Topic 06:** This topic appears to involve discussions about symbolism, religious sites, and potentially hidden meanings.

- **Topic 07:** This topic seems to cover discussions about technology, environmental phenomena, and potential hazards.

- **Topic 08:** This topic appears to be focused on economic discussions and financial systems.

- **Topic 09:** This topic revolves around religious and political discussions, particularly conflicts in the Middle East.

- **Topic 10:** This topic seems to involve broader discussions about the universe and space.

- **Topic 11:** This topic appears to blend discussions about geopolitics and environmental disasters, attempting to establish a link between the two subjects.

Table 4: Topics extracted by Top2Vec

| Topic | Relevant Keywords | Annotation |
|---|---|---|
| 1 | النساء، الرجال، الاجتماعي، المساواة، الطبيعة، الجنس، الزواج، النسوية، التواصل، المجتمع، القوانين، الشخصية، الحب، الأفكار، النفسية، العقل، العمل، الوعي، الجنسية، الحياة، القيمة، السياسية، السلطة English translation: Women, men, social, equality, nature, gender, marriage, feminism, communication, society, laws, personality, love, ideas, psychology, mind, work, awareness, sexuality, life, value, politics, power. | Social Equality and Gender Roles |
| 2 | الناتو، ليبيا، الاسرائيلية، العربي، العسكري، الأمريكي، المنطقة، سوريا، الايراني، واشنطن، قوات، الخليج، مضيق، حزب، تونس، افغانستان، استيراتيجية، الخارجية، صحيفة، الحرب، تقسيم، الصيهوني، مؤتمر، هرمز، مصر، المسلحة، قيادة، الحدود English translation: NATO, Libya, Israeli, Arab, military, American, region, Syria, Iranian, Washington, forces, Gulf, strait, party, Tunisia, Afghanistan, strategy, foreign affairs, newspaper, war, division, Zionist, conference, Hormuz, Egypt, armed, leadership, borders. | Geopolitical and Military Strategies |
| 3 | سطح، فضائية، كائنات، الاجسام، الفضاء، غريبة، كوكب، ناسا، الطائرة، الارض، جوية، وجود، رحلة، العلمي، الاطباق، مختلفة، علماء، المكان، تجربة، اكتشاف، موجودة، اتاركتيكا، النظرية، الازرق، البشر<br>English translation: Surface, space, creatures, bodies, space, alien, planet, NASA, aircraft, Earth, aerial, existence, journey, scientific, saucers, various, scientists, place, experiment, discovery, present, Antarctica, theory, blue, humans. | Extraterrestrial Phenomena |
| 4 | كروية، مسطحة، السماء، السبع، الارض، الشمس، الجبال، القران، الايات، ناسا، القمر، دليل، طبقات، تدور، الكرة، النار، الخلق، كوكب، الافق، الماء، سطح، النهار، الاقمار، عظيم<br>English translation: Spherical, flat, sky, the heavens, Earth, Sun, mountains, Quran, verses, NASA, Moon, evidence, layers, rotate, globe, fire, creation, planet, horizon, water, surface, daytime, moons, great. | Cosmology and Planetary Structure |
| 5 | رسول، اهل، الامام، النبي، حديث، المهدي، خروج، مكة، صحيح، الساعة، مريم، القران، الشام، يقتل، الاية، رحمة، الدنيا، جبل، رجل، القيامة، الصلاة، فتنة، بيت، المسلمون، عيسى، الملائكة<br>English translation: Messenger, people, Imam, Prophet, Hadith, Mahdi, departure, Mecca, authentic, the Hour, Mary, Quran, Syria, killed, verse, mercy, the world, mountain, man, Resurrection, prayer, tribulation, house, Muslims, Jesus, angels. | Eschatology and Prophecies |
| 6 | شعار، بابل، القدس، رمز، سليمان، الهيكل، داوود، المقدس، الاسود، فرسان، الشياطين، ابليس، الماسونيين، القديم، المنتظر، الصليب، العين، ملكة، السحر، الاقصى، المسيح، الاله، الانبياء، صهيون، الكتاب، الشيطان، اشارة، القصة، الشر، المتنورين، غريب<br>English translation: Logo, Babylon, Jerusalem, symbol, Solomon, temple, David, holy, black, knights, demons, Satan, Freemasons, ancient, awaited, cross, eye, queen, magic, Al-Aqsa, Christ, God, prophets, Zionist, book, devil, sign, story, evil, Illuminati, stranger. | Symbolism and Occultism |
| 7 | برنامج، الزلازل، استخدام، الطقس، سلاح، الارضية، التقنية، هارب، الغلاف، موجات، الكهرومغناطيسية، الدماغ، مشروع، التحكم، نووي<br>English translation: Program, earthquakes, use, weather, weapon, seismic, atmospheric, technology, HAARP, ionosphere, waves, electromagnetic, brain, project, control, nuclear. | Technology and Environmental Impact |
| 8 | الديون، الدولار، العملة، المالية، الصين، الدولي، اسعار، النقد، المالي، الاقتصاد، العملات، التجارة، النفط، الذهب، السوق، ازمة، انهيار، البنك، عالمي، الاوروبي، اسيا، الشركات<br>English translation: Debts, dollar, currency, financial, China, international, exchange rates, cash, monetary, economy, currencies, trade, oil, gold, market, crisis, collapse, bank, global, European, Asia, companies. | Economic and Financial Trends |
| 9 | الحكم، الشيعة، فوضى، الصهاينة، الثورات، القدس، الانهيار، الجمعة، المسجد، الاقصى، العربي، الاخوان، تنظيم، الحرب، الفساد، الفرات، تدمير، النيل، الاسلامية<br>English translation: Governance, Shiites, chaos, Zionists, revolutions, Jerusalem, collapse, Friday, mosque, Al-Aqsa, Arab, Muslim Brotherhood, organization, war, corruption, Euphrates, destruction, Nile, Islamic. | Religious and Political Conflicts |
| 10 | البحر، الفضائية، الكون، جسم، دخل، ناسا، رسالة، الكلام، ساعة، بشر، كائنات، الفضاء<br>English translation: The sea, space, the universe, body, income, NASA, message, speech, hour, humans, creatures, space. | Space and Cosmic Concepts |
| 11 | اعلان، عسكرية، حدوث، جهاز، قوة، سلاح، الناتو، ضرب، الروسي، نووي، تركيا، الزلزال، المخابرات، غريبة، طبيعية، الصواريخ، قاعدة، اليابان، هارب، امريكية، المياه، الموقع، عمليات، موجات، البحر، سوريا<br>English translation: Announcement, military, occurrence, device, force, weapon, NATO, strike, Russian, nuclear, Turkey, earthquake, intelligence, strange, natural, missiles, base, Japan, escapee, American, water, location, operations, waves, sea, Syria. | tensions, military strategies, and potential disasters |

# 4    Discussion

## 4.1    Arabic Conspiracism: Insights from Top2Vec Extracted Topics

While the Top2Vec method initially identified 11 topics, a thorough analysis of the results suggests that this count can be reduced to a total of 9 topics. This entails combining Topic 10 with Topic 4 and merging Topic 11 with Topic 7 due to the similarity in their keyword themes.

Based on these findings, the prevalent conspiracy theories discussed within the Arab virtual social landscape encompass the following categories: gender/feminist conspiracy theories, geopolitics conspiracy theories, government cover-ups related conspiracies, apocalyptic conspiracy theories, Judeo-Masonic conspiracy theories, and geoengineering conspiracy theories.

### 4.1.1    The Gender/Feminist Conspiracy

As per reference [41], gender conspiracy beliefs involve the idea that gender activists and advocates operate covertly to advance an agenda conflicting with established societal norms. This viewpoint portrays them, including feminists and LGBT+ proponents, as a malevolent collective, criticized for potentially causing gender conflicts, undermining parenthood, and eventually dismantling the cherished family [34] — an essential principle upheld by conservative societies, as exemplified by Arab societies.

Conspiracy theories concerning gender ideology and feminism in the Arab context assert that these movements are propagating Western cultural values that jeopardize Arab society. We posit that this subject prominently dominates discussions within online Arab conspiracist communities, possibly due to the visibility of feminism and LGBT+ activism on the Web and social media since Arab Spring uprisings [42][43]. Coupled with the Westernized nature of these two movements, this renders their messages not only somewhat irrelevant but also conceptually aggressive to a significant extent, making them the primary target of conspiratorial narratives [44].

Additional justification for the prominence of gender conspiracies within Arabic online content is evident when considering the context of LGBT+ rights situation in the Arab world. In fact, LGBT+ matters encounter substantial resistance from both individuals and many governments in the Middle East and North Africa, which firmly reject the concepts of 'sexual orientation' and 'gender identity' altogether [45]. With some of the world's most restrictive legislation for LGBT+ individuals, the region is also home to five out of the 11 UN-member states that have prohibited the death penalty for consensual same-sex relations [46]. As for feminism, perceptions are intricate and divergent, varying from outright rejection to cautious acceptance. However, a prevailing view among Arabs is that feminism represents an alien Western-imported concept incompatible with Islamic values and conservative Oriental societal norms [44]. This viewpoint holds sway across both men and women, with particular potency in the more conservative segments of society [47].

### 4.1.2    Geopolitical Conspiracies

Geopolitical conspiracy theories focus on the actions and motivations of governments, organizations, and other actors on the global stage. These theories often involve allegations of hidden, manipulative, or secret plots operating within global or regional political dynamics. It often involves the idea that certain events are orchestrated or influenced by secretive organizations, powerful nations or elites, or shadowy forces seeking to achieve specific goals that are hidden from public view.

Geopolitical conspiracy theories can be used to explain complex political events or situations. In our corpus, they encompass a wide range of topics, including international relations and power struggles (Topic 2), economic and financial issues (Topic 8), as well as religious and political regional conflicts (Topic 9).

It's important to acknowledge that certain geopolitical conspiracy theories might be rooted in valid concerns surrounding transparency, corruption, or the misuse of power [19][48]. Scholarly investigations revealed that the encounter of Arab region with colonialism and the establishment of Israel has fostered a persistent sense of suspicion regarding foreign motives and intentions [19]. Furthermore, a substantial negative sentiment towards the United States prevails across the region. This sentiment is rooted in the history of U.S. involvement in the area, particularly evident since the inception of the War on Terror [49]. Consequently, the prevalence of belief in conspiracy theories regarding the Western world, Jewish communities, and Israel is considerable. These beliefs are closely intertwined with overarching anti-Western and anti-Jewish attitudes, particularly noticeable among individuals possessing a high level of political awareness [21].

Within the same framework, Middle Eastern conspiracy theories concerning religion and its resulting political conflicts are frequently employed to shift accountability away from the indigenous origins of political and religious sectarianism, often depicted as a Western influence aimed at fragmenting the Arab Muslim world [50]. Likewise, economic manipulation conspiracy theories are also prevalent in this context. Despite being endowed with abundant natural and human resources, Arab nations frequently confront a range of economic and financial obstacles that significantly influence the future possibilities and overall welfare of their populations. These challenges encompass high unemployment rates, the economic vulnerability stemming from oil price fluctuations (due to oil dependence), unequal distribution of income, issues of corruption, and the impact of political instability [51].

Conspiratorial narratives condemn the external efforts of influential nations (primarily the USA) and international entities (such as the IMF and EU) for aiming to economically destabilize the region and incite conflicts to exploit its resources. These conspiracy theories extend beyond this restricted viewpoint; they also encompass the global economy, propagating conspiratorial narratives about international economic crises and foreseeing potential economic collapses as outcomes of concealed plots and agendas.

### 4.1.3 Government Cover-ups related Conspiracies

Conspiracy theories related to this topic might involve claims of government cover-ups about the true nature of the Earth, existence of alien life, secret technology reverse-engineered from crashed UFOs, or hidden interactions between humans and extraterrestrial beings.

- **UFO**: Originating as a distinctly American occurrence, the contemporary fascination with Unidentified Flying Objects (UFOs) can be dated back to 1947 [52], marked by the initial surge of UFO sightings. Discussions surrounding the potential existence of extraterrestrial life have persisted throughout the ensuing decades, attaining unparalleled prominence.

  As referenced in [53], what started as an American innovation quickly evolved into a global sensation, particularly gaining traction in Western Europe. Subsequently, this phenomenon proliferated to various corners of the globe, facilitated by globalization, the internet, and social media. While there is an absence of literature concerning UFO sightings in Arab countries, the authors in [53] point to North Africa, among other regions, as a witness to this phenomenon, with the first sightings recorded in 1950. However, it's important to note that many people around the world, particularly in the Arab world, hold reservations regarding the authenticity of these occurrences. Drawing upon the temporal analysis of the data presented in Section 3, the findings reveal the relatively recent emergence of this subject within Arabic online conspiracism. Our assertion proposes that discussions about extraterrestrial existence serve no purpose other than to capture the attention of followers and align with ongoing global discussions on this phenomenon. This interest resurged in public discourse following reports from February 2023, which detailed the U.S. military's confirmed interception of mysterious flying objects (later identified as Chinese surveillance balloons) across the skies of both the United States and Canada [54].

- **Flat Earth**: The Flat Earth theory is a belief that contradicts the widely accepted scientific understanding of the Earth's shape, which is an oblate spheroid. Flat Earthers claim that the Earth is not a spherical object but rather a flat, disc-shaped plane.

  The concept of the Flat Earth originated from a literal Biblical interpretation as a form of science denial [55] [32][56]. The modern Flat Earth ideology emerged in 1849 in England as a fundamentalist religious response against the scientific consensus on Earth's shape. This perspective gained traction with the formation of the International Flat Earth Research Society of America (commonly known as the Flat Earth Society) in California, USA, in 1971 [57].

  Scholarly investigations have unveiled that the substantial engagement of Flat Earthers on YouTube has led to the widespread dissemination of their beliefs [55] [58]. A study conducted in 2022 [59] indicated that as of July 2021, approximately four million individuals had subscribed to the top 122 YouTube channels dedicated exclusively to sharing flat Earth content. This influence has even extended to the Arab online sphere, sparking debates regarding the shape of the Earth, which has long been regarded as spherical within Muslim Arab literature. Unlike Christian traditions, Islamic scholars and theologians have historically reached a consensus on the Earth's spherical nature for many centuries [60]. This consensus was established based on the contributions of geography and astronomy scholars during the Golden Age of Arab civilization in the medieval era. These studies found validation in the Quran, which distinctly denotes the Earth's spherical form in various verses [60]. Nevertheless, proponents of the Flat Earth movement within the Arab sphere persistently endeavor to substantiate their convictions by delving into unconventional interpretations of Quranic verses related to the cosmos. This often entails adopting a literal interpretive approach for sacred texts, akin to the approach embraced by their Western counterparts. Additionally, it involves utilizing pseudo-science and raising uncertainties about the reliability and credibility of scientific institutions through allegations of cover-ups, presented as conspiracies.

### 4.1.4 Apocalypticism

In its broader academic context, the term 'apocalypse' is employed to denote the conviction regarding an impending significant event, typically of cataclysmic magnitude or epochal transformation, for which a limited number of individuals possess prior knowledge, enabling them to undertake suitable preparations [61].

Apocalyptic beliefs, whether religious or secular, are deeply rooted in religious, cultural, or philosophical traditions. They encompass a diverse range of interpretations concerning catastrophic events, often with the potential to bring about world-ending scenarios. These beliefs often share commonalities with conspiracy theories, including predictions about future events, a prevailing sense of impending catastrophe, and the belief in concealed knowledge or hidden forces shaping the world's destiny [62] [63].

Since World War II, interest in apocalyptic prophecies has seen a resurgence in modern societies, including Arab countries, driven by two primary factors. The development and widespread availability of nuclear weapons have instilled a lasting fear of humanity's potential for self-destruction [64]. Additionally, sociological changes, including globalization, individualization, cosmopolitanism, detraditionalization, and digital technology expansion, have ushered in a noteworthy period of historical transition marked by significant uncertainty [62].

In the Arab context, a history of political instability, recurring conflicts, and regional upheavals spanning many centuries has sustained a persistent interest in End-Time prophecies. This environment often leads to the intersection of apocalyptic religious beliefs and conspiracy theories, where religious narratives occasionally form the basis for or influence conspiracy theories [18]. Conversely, a conspiratorial mindset can prompt individuals to reinterpret religious prophecies within a more conspiratorial or apocalyptic framework.

### 4.1.5 The Judeo-Masonic Conspiracy Theory

The Judeo-Masonic conspiracy theory is a political ideology that claims that a secret alliance of Jews and Freemasons is plotting to control global events and institutions for their own nefarious purposes. The theory emerged in Europe in the late 18th and early 19th centuries. It was instrumentalized by various conservative and reactionary groups to oppose the French Revolution, as well as modern ideologies like liberalism, democracy, socialism, communism, and other progressive movements. The theory also blamed Jews and Freemasons for causing wars, revolutions, economic crises, social unrest, and moral decay [65].

The prevalent theme of the Judeo-Masonic myth is notably pervasive within Middle Eastern offline and online conspiratorial narratives. However, it is important to recognize that this myth, while now prominent in the Arab world, finds its origins as an 'Islamization of European Antisemitism,' as asserted by [66].

Initially disseminated within Catholic Arab communities, the narrative of the Judeo-Masonic conspiracy gradually permeated Islamic circles, gaining traction only in the 1920s due to escalating tensions between Jews and Arabs in Palestine [67]. Subsequently, it evolved into a central tenet of opposition against Zionism following the establishment of the state of Israel, culminating in its zenith during the 1980s and 1990s [67]. Throughout this evolution, religious institutions, initially within the Catholic framework and later within the Islamic sphere, played a significant role in amplifying anti-Jewish and anti-Masonic sentiments [67].

### 4.1.6 Geo-engineering Conspiracy Theories

Geo-engineering conspiracy theories revolve around speculative and often unfounded beliefs that powerful entities, such as governments or corporations, are clandestinely engaged in large-scale manipulation of the Earth's environment. These theories propose that these entities are deliberately altering natural processes for their own hidden agendas, often with the assumption that the public is being kept unaware or deceived about these activities [68]. Common themes within geo-engineering conspiracy theories include: climate control and modification [69], weather manipulation [70], earthquake and natural disaster induction [71], chemtrails and atmospheric aerosols [72], the High-frequency Active Auroral Research Program (HAARP) [73], and population control [74].

The keywords within Topic 07 (refer to Table 6) show that earthquakes, the HAARP project, and weather are central subjects of discussion. This focus becomes further apparent upon examining the keywords within Topic 11, which specifically addresses the recent earthquake in Turkey and Syria.

Earthquakes occur suddenly and unexpectedly, resulting in significant and, at times, devastating social and economic repercussions. Given this reality, they have the potential to evoke anxiety, fear, and even panic in individuals, rendering them susceptible to embracing misinformation [71].

The earthquake that struck Turkey and Syria in February 2023 not only brought about physical devastation but also sparked a surge in conspiracy theories on the Arab sphere of social media. These theories sought to attribute the earthquake's cause to various entities and technologies, fueling speculation and mistrust [75]. One theory implicated the United States, suggesting they used a nuclear bomb or secret weapons to induce the earthquake, potentially tied to the HAARP project. Another proposed high-density electromagnetic pulse weapons as the cause. A different narrative focused on Turkey's dams, suggesting they led to groundwater leakage and the earthquake. A fourth theory linked the earthquake to a failed Russian-Turkish nuclear experiment at the Akkuyu Nuclear Power Plant, meant for electricity generation but resulting in seismic disturbances.

## 4.2 Research Implications

Integrating insights from data analysis, Named Entity extraction, and Topic Modeling, we developed a comprehensive understanding of Arabic conspiracist content on the web and social media.

Our temporal analysis uncovered the evolving interests of Arab conspiracists over time. Currently, the most prominent themes encompass society, specifically societal roles, pseudo-science, and to some extent, religion. This signifies a notable departure from the previous predominance of geopolitical themes, which held sway from the early 2010s until 2017.

The Named Entity extraction method has provided a holistic view of the key entities shaping Arabic conspiracist discussions. Noteworthy is its successful identification of entities that are related to relevant topics such as geopolitics (encompassing economic concerns and political conflicts), religion, and catastrophic events. However, it fell short in capturing discussions pertaining to societal debates. This limitation is explicable, given that discussions around societal roles often delve into abstract socio-cultural and ideological concepts that diverge from the typical patterns of named entities.

Topic modeling addressed this gap and facilitated a deeper understanding of the fundamental beliefs driving these conspiracist discussions. Top2Vec contextualized the extracted named entities, facilitating a more nuanced understanding of the conspiracist narrative. The investigation of its findings revealed that Arabic online conspiracism revolves around six major conspiracy theories: gender/feminist conspiracy theories, geopolitical conspiracy theories, government cover-ups, apocalypticism, Judeo-Masonic conspiracy theories, and geoengineering conspiracy theories. Our in-depth examination of prior studies and literature showed that some theories are deeply rooted in Arab culture, particularly those related to geopolitics, apocalypse scenarios, and clandestine societies. This category of theories continues to gain traction amid the ongoing political and economic instability in the Arab region. Others have more recently captured the attention of Arab conspiracists, such as gender/feminist conspiracy theories, emerging in response to the noticeable online activism of gender/feminist movements in recent years. Lastly, a third category of theories, including geoengineering, the flat earth theory, and UFOs, emerges sporadically in response to diverse triggers, such as natural disasters, scientific denial, and unexplained aerial phenomena or sightings of unidentified flying objects.

### 4.3 Theoretical Implications

Topic Modeling and Named Entity Recognition (NER) represent effective methodologies for comprehending Arabic online conspiracist content and monitoring discussions related to conspiracy topics. Their synergistic application efficiently delves into textual data, producing valuable insights.

In addition, the study highlights the pivotal role of cultural factors in modeling conspiracy beliefs, implying the need to consider the correlation between socio-political instability and the appeal of conspiracy ideologies. Furthermore, it posits that the dissemination of conspiracy theories can be construed as a manifestation of social resistance or a form of protest against perceived threats or injustices prevailing within the Arab region's sociopolitical landscape. The study also accentuates the adaptability of conspiracist narratives to contemporary social and political developments, emphasizing the necessity to apprehend the triggers underpinning their sporadic emergence.

### 4.4 Practical Implications

The proliferation of conspiracy theories in the Arab online landscape presents a multifaceted challenge that lacks straightforward solutions. Nevertheless, this study serves as a crucial initial step toward comprehending this intricate phenomenon and formulating effective countermeasures.

The study's findings offer valuable insights for a diverse array of stakeholders, encompassing social media platforms, government agencies, and researchers. These insights enable more informed responses to online conspiracist content, thereby informing the development of multifaceted strategies. Such strategies may include various approaches, such as heightening awareness regarding the perils associated with conspiracy theories and facilitating their detection.

Moreover, these strategies may involve the promotion of critical thinking skills and media literacy among the general populace. Additionally, they may extend to providing support for fact-checking initiatives and endeavors aimed at debunking false information, thereby enhancing the public's capacity to discern accurate information from misinformation.

Simultaneously, it remains imperative to address the root causes underpinning conspiracy theories, including factors like political instability and economic adversity, as part of a comprehensive strategy aimed at mitigating their impact on society.

## 5 Conclusion

The dissemination of misinformation within the Arabic-speaking online sphere represents a significant concern that warrants serious attention. In light of this, our research integrates temporal analysis, Named Entity Recognition, and Topic Modeling to unveil conspiracist narratives in Arabic online content, analyze their characteristics and origins, and offer valuable insights into their dynamics. Our comprehensive analysis uncovers an evolving narrative within Arabic conspiracy theories. This evolution is characterized by temporal variations in dominant themes, a pronounced focus on geopolitical dynamics and influential entities, and a diverse range of thematic concerns. The most prominent conspiracies that have captured the attention of conspiracy theorists in the Arab-speaking online communities encompass gender/feminist conspiracies, geopolitical conspiracies, government cover-ups, apocalyptic scenarios, Judeo-Masonic conspiracies, and geoengineering conspiracies. The examination of the underlying roots of these conspiracy theories underscores the critical importance of recognizing the influence of social, cultural, and historical factors when addressing conspiracism in the Arabic context. Subsequent research endeavors should be dedicated to untangling the intricate web of these factors and validating the kinds of conclusions that can be drawn from this study.

## 6 Limitations and Future Work

Although our study has contributed to our understanding of conspiracy thinking in Arab societies, it is essential to acknowledge its limitations. One notable limitation is the inherent constraint of topic models and NER methodology in fully capturing the multifaceted nature of the data. While these techniques have proven useful, they may not encompass the full breadth and depth of the conspiracist online discourse. Moreover, our analysis, while revealing prevalent conspiracy theories, has only scratched the surface of a complex phenomenon. We recognize that a more comprehensive exploration of the psychological and sociological factors underpinning individuals' belief in these theories is necessary.

To enhance the accuracy and scope of our analysis, we recommend refining Named Entity extraction and Topic Modeling algorithms designed specifically to accommodate the complexities of the Arabic language and culture. This improvement promises to augment our ability to capture a broader range of discussions and detect emerging trends. Complementing this linguistic refinement, a detailed examination of the linguistic features within conspiracy discourse is essential. Analyzing elements such as word choice, sentence structure, and rhetorical strategies used by conspiracists will provide deeper insights into the construction of persuasive arguments and how individuals are swayed to embrace these theories. Beyond the Arabic-speaking world, there is a compelling need for cross-cultural comparisons. Such analyses would facilitate the identification of both commonalities and disparities in the formulation, dissemination, and reception of conspiracy theories globally. This cross-cultural perspective is invaluable for understanding universal patterns and regional variations. Furthermore, investigating the impact of conspiracist narratives on public opinion and behavior remains a crucial research frontier. Understanding the influence of these beliefs on decision-making processes and societal attitudes is pivotal in assessing potential societal implications and devising strategies to counteract any detrimental effects.